\RequirePackage{fix-cm}
\documentclass[twocolumn]{svjour3}          % twocolumn
\smartqed  % flush right qed marks, e.g. at end of proof
\usepackage{graphicx}
\usepackage[numbers]{natbib}
\usepackage{multirow}
\usepackage{amsmath}
\usepackage{amssymb}
\usepackage{bbm}
\usepackage{bm}
\usepackage{booktabs}
\usepackage{array} % for extended column definitions, e.g. >{\em}c
\usepackage{blindtext}
\usepackage{comment}

\usepackage{xspace}

\makeatletter
\DeclareRobustCommand\onedot{\futurelet\@let@token\@onedot}
\def\@onedot{\ifx\@let@token.\else.\null\fi\xspace}

\def\eg{\emph{e.g}\onedot} 
\def\ie{\emph{i.e}\onedot}

\makeatother

\usepackage{pifont}

\usepackage[dvipsnames]{xcolor}
\usepackage{color, colortbl}
\definecolor{citecolor}{HTML}{0071bc}
\definecolor{tabhighlight}{HTML}{e5e5e5}
\usepackage[colorlinks,citecolor=citecolor]{hyperref}
\makeatletter
\renewcommand\paragraph{
  \@startsection{paragraph} % name
  {4} % level
  {\z@} % indent
  {.5em \@plus1ex \@minus.2ex} % beforeskip
  {-.5em} % afterskip
  {\normalfont\normalsize\bfseries} % style
}
\makeatother
\begin{document}
\sloppy

\title{ReliTalk: Relightable Talking Portrait Generation from a Single Video %\thanks{Grants or other notes
}


\author{Haonan Qiu  \and
        Zhaoxi Chen \and
        Yuming Jiang \and
        Hang Zhou \\ \and
        Xiangyu Fan \and
        Lei Yang \and
        Wayne Wu \and
        Ziwei Liu
}


\institute{Haonan Qiu \at
              S-Lab, Nanyang Technological University, Singapore \\
              \email{HAONAN002@e.ntu.edu.sg}
           \and
           Zhaoxi Chen \at
              S-Lab, Nanyang Technological University, Singapore \\
              \email{ZHAOXI001@e.ntu.edu.sg}
           \and
           Yuming Jiang \at
              S-Lab, Nanyang Technological University, Singapore \\
              \email{YUMING002@e.ntu.edu.sg}
           \and
           Hang Zhou \at
              The Chinese University of Hong Kong, China \\
              \email{zhouhang@link.cuhk.edu.hk}
           \and
           Xiangyu Fan \at
              SenseTime Research, China \\
              \email{fanxy1993@gmail.com}
           \and
           Lei Yang \at
              SenseTime Research, China \\
              \email{yanglei@sensetime.com}
           \and
           Wayne Wu \at
              SenseTime Research, China \\
              \email{wuwenyan0503@gmail.com}
           \and
           Ziwei Liu \at
           S-Lab, Nanyang Technological University, Singapore \\
           \email{ziwei.liu@ntu.edu.sg}
}

\date{Received: date / Accepted: date}

\maketitle

\begin{abstract}
    Recent years have witnessed great progress in creating vivid audio-driven portraits from monocular videos. However, how to seamlessly adapt the created video avatars to other scenarios with different backgrounds and lighting conditions remains unsolved. On the other hand, existing relighting studies mostly rely on dynamically lighted or multi-view data, which are too expensive for creating video portraits.
    To bridge this gap, we propose \textbf{ReliTalk}, a novel framework for relightable audio-driven talking portrait generation from monocular videos.
    Our key insight is to decompose the portrait's reflectance from implicitly learned audio-driven facial normals and images.
    Specifically, 
    we involve 3D facial priors derived from audio features to predict delicate normal maps through implicit functions. These initially predicted normals then take a crucial part in reflectance decomposition by dynamically estimating the lighting condition of the given video.
    Moreover, the stereoscopic face representation is refined using the identity-consistent loss under simulated multiple lighting conditions, addressing the ill-posed problem caused by limited views available from a single monocular video.
    Extensive experiments validate the superiority of our proposed framework on both real and synthetic datasets. 
    Our code is released in (\href{https://github.com/arthur-qiu/ReliTalk}{https://github.com/arthur-qiu/ReliTalk}).
\end{abstract}


\begin{figure*}
    \centering
    \includegraphics[width=0.99\textwidth]{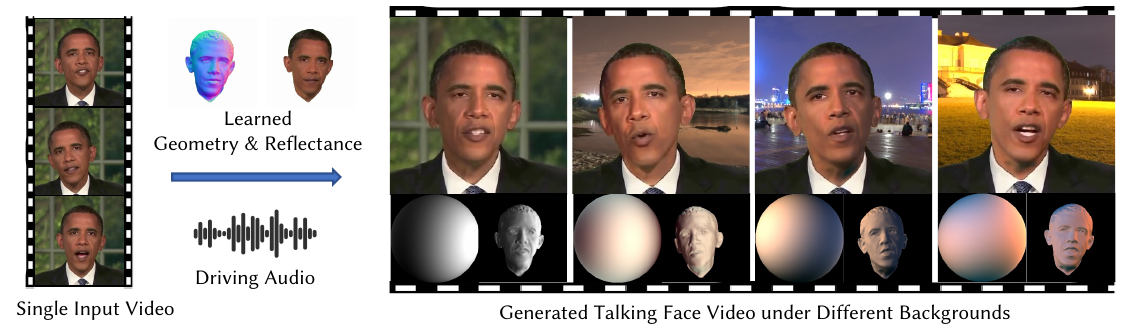}
    % \vspace{-0pt}
    \caption{\textbf{Relighting Talking Portrait with Assigned Background.} (Left) Our method takes a monocular video as input and estimates the corresponding normal and albedo which can be driven by audio. (Right) Talking portrait renderings with different illuminations, where lighting and shading are placed at the bottom. The rightmost three are relighted by HDR background images. Only a single video is required as the training data, without any extra annotations.}
    \label{fig:teaser}
\end{figure*}%

\section{Introduction}

Creating personalized audio-driven talking portraits has many applications in teleconferencing, video production, VR/AR games, and the movie industry. 
Given its great potential, research on talking face generation~\cite{taylor2017deep, thies2020neural, zhou2019talking, zhou2021pose, zhang2021flow, ji2021audio} has enjoyed massive popularity in recent years, with emphasis on creating lip-synced~\cite{prajwal2020lip,thies2020neural} portraits with diverse head motions, talking styles, and emotions~\cite{yi2020audio,wu2021imitating}. 
However, the ability to change the lighting conditions of audio-driven portraits is still under-explored, which is critical to real-world applications as we expect the portrait in the foreground to be seamlessly harmonized with backgrounds under different illuminations.

To generate a relightable talking portrait from a single video, we argue that the underlying model should be capable of \textbf{1)} estimating fine-grained 3D head geometry from monocular videos, \textbf{2)} reflectance decomposition without any extra annotations, and \textbf{3)} generalizing to driven audios. 
However,
most learning-based methods either operate only on the 2D plane~\cite{zhou2019talking, zhou2021pose, prajwal2020lip}, or leverage structural intermediate representations~\cite{chen2019hierarchical, cudeiro2019capture, ji2021audio,thies2020neural,wu2021imitating,zhou2020makelttalk},
and neural radiance fields~\cite{guo2021ad, yao_dfa-nerf_2022, shen2022dfrf}. No fine-grained 3D geometry can be acquired for reflectance decomposition in these studies.
On the other hand, adapting existing relighting techniques~\cite{sun2019single, wang2020single, pandey2021total, zhang2021neural} is too expensive for audio-driven video portraits given their dependence on multi-view or dynamically lighted data.

To bridge this gap, we propose \textbf{ReliTalk}, a novel framework for relightable audio-driven talking portrait generation that only requires a single monocular video as input, as shown in Fig.~\ref{fig:teaser}. Our key insight is the self-supervised implicit decomposition of geometry and reflectance, both of which can be further driven by input audios.
In specific, the proposed approach first extracts expression- and pose-related representations based on 3D facial priors~\cite{li2017learning}, and refines them into delicate normal maps through implicit functions. The initial normals then take a critical role in reflectance decomposition, which disentangles the human head as a set of intrinsic normal, albedo, diffuse and specular maps, by dynamically estimating the lighting condition of the given video.
To get rid of leveraging knowledge from expensive capturing data (\ie Light Stage~\cite{lightstage}), we carefully design several learning objectives to decompose the human portrait into corresponding maps from monocular videos, which will be introduced in the following sections.

To learn the audio-to-face mapping that better generalizes to unseen audio, we introduce mesh-aware guidance to assist the lip-syncing especially when the training video is too short to cover enough audio variance. Specifically, we use a model pre-trained on the VOCA dataset~\cite{8954000} to obtain lip-related meshes as the additional guidance. 
Phoneme-related features and lip-related meshes are separately encoded and then concatenated to achieve more accurate audio-driven animations. 
Phoneme-related features enable the network to learn richer mouth shapes and mesh-aware features provide coarse information on the opening and closing of lips, even if the input audio is far away from the audio used in training. 

Natural talking portrait videos usually provide a limited perspective of the target persons when they face the camera without turning around. Plus, the lack of multi-view information inherently negatively impacts an accurate estimation of 3D geometry. To address the ill-posed inverse problem of geometry and reflectance decomposition caused by single-view, limited motion variance, and unknown illuminations, we design identity-consistent supervision (ICS) with simulated multiple lighting conditions to refine normal maps. The key insight is that we relight the human portrait on-the-fly during the training stage, by sampling different lights and using identity-consistent loss to update normal maps. 

\begin{figure*} %[t]
% \vspace{-0.1in}
\centering
\includegraphics[width=0.90\linewidth]{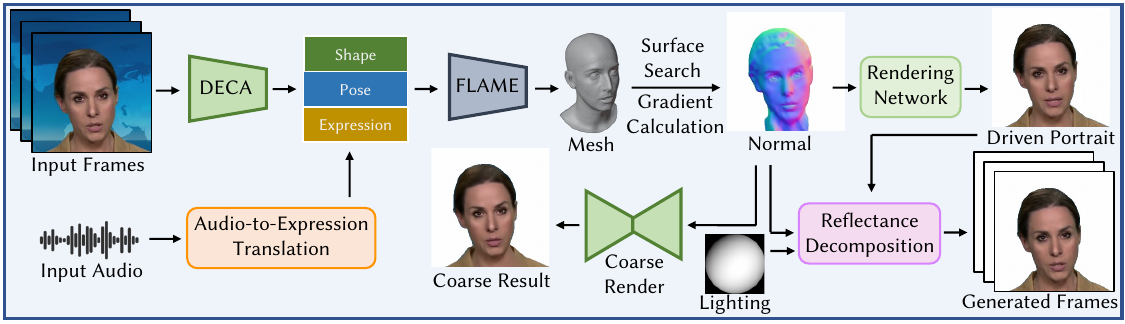}
\vspace{-0pt}
\caption{\textbf{Overview of Our Proposed Framework.} Denote the input video with unknown illuminations as $\textbf{V} = \{I_{1}, I_{2}, ..., I_{t}\}$ with audio sequence $\textbf{a} = \{a_{1}, a_{2}, ..., a_{t}\}$, where $t$ is the number of frames. Generally, our aim is to extract the geometry and reflectance information from video $\textbf{V}$ in an unsupervised manner then drive the geometry deformation according to the audio. 
}
\vspace{-0pt}
\label{fig:pipeline}
\end{figure*}

We evaluate our approach on both real and synthetic datasets. Overall, \textbf{ReliTalk} drives and relights dynamic human portraits in high fidelity, outperforming other methods on both perceptual quality and reconstruction correctness.
Our contributions are summarized as follows:

\begin{itemize}
    \item We propose a novel framework \textbf{ReliTalk} that learns relightable audio-driven talking portrait generation and only requires a single monocular portrait video.
    \item We propose the additional audio-to-mesh guidance to improve the mapping accuracy especially when the single training video only has a limited audio variance.
    \item We design identity-consistent supervision with simulated multiple lighting conditions, addressing the ill-pose problem caused by limited views available from the single video.
\end{itemize}

\section{Related Work}

\noindent\textbf{Inverse Rendering.}
Recovering and disentangling the appearance of observed images into geometry and reflectance is a long-standing problem in the field of computer vision and graphics. Prior works~\cite{barron2014shape, intrinsic} address this challenge by physical-based priors on synthetic image data. However, they fail to extract the underlying 3D representation. 
Later approaches~\cite{chan2022efficient, or2022stylesdf, xu20223d, sun2022fenerf, zhao2022generative, pan20202d, chan2021pi} successfully extract the 3D representations by the 3D generator and refine the output using image-based CNN networks.
Recently, methods based on implicit representation~\cite{zhang2021nerfactor, nerv2021} propose learning 3D reflectance and geometry from multi-view images. In this work, we aim to tackle a harder problem, \ie, inverse rendering from a monocular video of a talking human face. Note that, limited-view information can be accessed as the person is always oriented toward the front. 

\noindent\textbf{Portrait Relighting.}
One-Light-at-A-Time (OLAT) capturing system allows for obtaining detailed portrait geometry and reflectance. Many methods based on it have achieved impressive success~\cite{sun2019single, wang2020single, pandey2021total, zhang2021neural}. However, it is only applicable in a constrained environment due to its complexity and expense. 
Other methods~\cite{zhou2019deep, hou2021towards, hou2022face, caselles2023sira} simulate some multi-lighting data and train the network to predict relighted results. Due to their limited simulation methods, the final results are far away from OLAT-based methods. 
\cite{yeh2022learning} synthesizes a high-quality multi-lighting dataset but it is still not available to the public. Another simplified strategy that requires the user to capture a selfie video or a sequence of images to gain multi-view information is proposed~\cite{nestmeyer2020learning, wang2022sunstage}. 
And Relighting4D~\cite{chen2022relighting4d} can even relight dynamic humans with free viewpoints only from videos. However, their rendering quality is totally tied to the accuracy of geometry, requiring enough viewpoints from videos. 
Our method is able to relight portraits with finer details from the monocular portrait video even without much multi-view information available.

\noindent\textbf{Audio-driven Talking Face.}
Face animation has wild applications, drawing great research interest in computer vision and graphics. Recent methods for audio-driven animation \cite{cudeiro2019capture, faceformer2022, 10.1145/3072959.3073658, 9710491,suwajanakorn2017synthesizing, kim2018deep} are usually data-driven and can be divided into two categories. One is generalized animation~\cite{cudeiro2019capture, faceformer2022, 9710491}, which utilizes some large datasets which contain the pair data of audio/speech to lip/face. Wave2Lip~\cite{prajwal2020lip} trains a mapping from audio to lips on LRS2~\cite{Chung17}. Instead of learning a highly heterogeneous and nonlinear mapping from audio to video directly, Everybody’s Talkin~\cite{song2022everybody} additionally involves the statistical linear 3D face model and builds an easier map from audio to parameters of 3DMM~\cite{blanz1999morphable}. 
Our proposed method takes a similar strategy that drives the whole portrait through controlling the parameters of FLAME model~\cite{li2017learning}.
The other one is personalized animation~\cite{suwajanakorn2017synthesizing, 10.1145/3072959.3073658, tang2022real}, which usually does not rely on a large dataset for training and only builds one model for each person. Recently, with the emergence of Neural Radiance Fields (NeRF)~\cite{mildenhall_nerf_2020, barron_mip-nerf_2021, barron_mip-nerf_2021-1}, many NeRF-based audio-driven methods are proposed~\cite{guo2021ad, liu2022semantic, yao_dfa-nerf_2022}. However, those methods can not drive the portraits well when meeting novel audio. DFRF~\cite{shen2022dfrf} improves this issue with a pre-trained base model but the final results are still not satisfactory. 

\section{Our Approach}
\label{sec:method}

\begin{figure*} %[t]
% \vspace{-0.1in}
\centering
\includegraphics[width=0.90\linewidth]{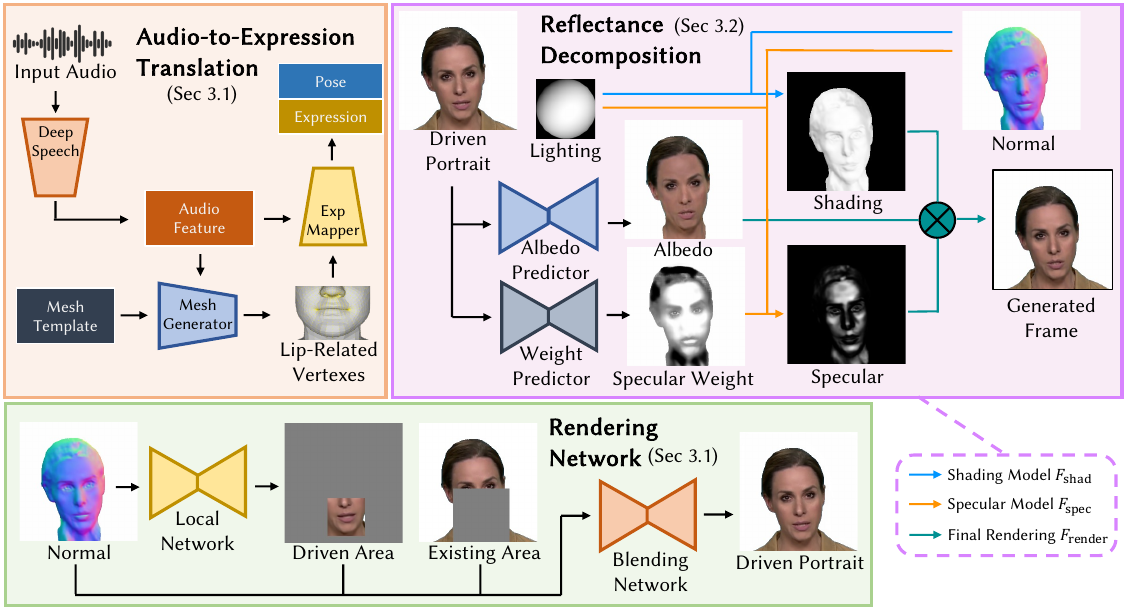}
\vspace{-0pt}
\caption{\textbf{Details of Our Proposed Framework.} 
Our framework decomposes the video $I$ into a set of normal $N$, albedo $A$, shading $S_\text{shad}$, and specular $S_\text{spec}$ maps. Specifically, we neurally model the expression- and pose-related geometry of human heads based on the FLAME model~\cite{li2017learning}. Then, the reflectance components are decomposed via multiple carefully designed priors (Section~\ref{sec:reflectance}).
With the well-disentangled geometry and reflectance, we use audio from the user to drive the human portrait by controlling expression and pose coefficients, then render it with any desired illuminations, which seamlessly harmonizes with the background.
}
\vspace{-0pt}
\label{fig:details}
\end{figure*}

Given a monocular video of a talking portrait, our framework can re-render the human portrait with novel illuminations driven by the input audio. Denote the input video with unknown illuminations as $\textbf{V} = \{I_{1}, I_{2}, ..., I_{t}\}$ with audio sequence $\textbf{a} = \{a_{1}, a_{2}, ..., a_{t}\}$, where $t$ is the number of frames. The key aim of our framework is to extract the geometry and reflectance information from video $\textbf{V}$ in an unsupervised manner, and the geometry deformation is driven by the audio accordingly.
Specifically, we neurally model the expression- and pose-related geometry of human heads based on the FLAME model~\cite{li2017learning}. Then, an audio-to-geometry mapping is learned to drive the portrait and also provide a good initial normal estimation (Section~\ref{sec:audio}). Meanwhile, the reflectance components, \textit{i.e.,} normal $N$, albedo $A$, shading $S_{shad}$, and specular $S_{spec}$ maps, are decomposed via carefully designed priors (Section~\ref{sec:reflectance}). During training, the lighting condition $L$ of the given video is estimated on-the-fly, and the training objective is reconstructing the whole video. 
In addition, multiple lighting conditions are randomly simulated for identity-consistent supervision which further refines geometry estimation.
With the well-disentangled geometry and reflectance, we use audio from the user to drive the portrait by controlling expression and pose coefficients, then render it with any desired illuminations, which seamlessly harmonizes with the background.
The whole pipeline is shown in Fig.~\ref{fig:pipeline}.

\subsection{Audio-Driven Synthesis}
\label{sec:audio}
\noindent\textbf{Expression- and Pose-related Geometry.} Estimating the surface normal of talking portraits from monocular videos is a non-trivial task, given the ill-posed nature of single-view reconstruction. To address this issue, we leverage a parametric model, FLAME~\cite{li2017learning}, as the human head prior to modeling the expression- and pose-related human portrait:
\begin{equation}
\text{FLAME}({\beta}, {\theta}, {\psi}): \mathbb{R}^{|{\beta}| \times|{\theta}| \times|{\psi}|} \rightarrow \mathbb{R}^{n\times 3},
\label{eq:flame}
\end{equation}
which takes coefficients of shape ${\beta} \in \mathbb{R}^{|{\beta}|}$, pose ${\theta} \in \mathbb{R}^{|{\theta}|}$, and expression ${\psi} \in \mathbb{R}^{|{\psi}|}$ as input. We use the off-the-shelf tool~\cite{feng2021learning} to estimate those parameters. Intuitively, this parametric human portrait model offers a good initialization of the 3D geometry, which facilitates further refinement.

However, this initial parametric portrait model is not well-aligned with the details of the given human portrait. To refine the initial model, we use nearest surface intersection search~\cite{zheng2022avatar} to optimize the initial mesh and calculate the normal $N$ as the normalized gradient on the surface. This normal $N$ will be further optimized during the reflectance decomposition process (Section~\ref{sec:reflectance}).

\noindent\textbf{Mesh-Aware Audio-to-Expression Translation.}
From the perspective of the mapping function, learning a direct mapping from audio to talking video is hard due to its high-dimensional property. In contrast, mapping audio signals to expressions and poses of the head is much easier. To enable robust talking portrait generation, we leverage a mesh-aware audio-to-expression translation strategy. Benefiting from FLAME~\cite{li2017learning} based design, our extracted head geometry is expression- and pose-related. We first use DeepSpeech~\cite{amodei2016deep} to extract phoneme-related audio features $f_\text{pho}$:
\begin{equation}
f_\text{pho} = \text{DeepSpeech}(a).
\end{equation}

Then extracted audio features $f_\text{pho}$ are fed to a model pre-trained on the VOCA dataset~\cite{8954000} to predict lip-related mesh vertices $V_\text{lip}$ as the additional guidance:
\begin{equation}
V_\text{lip} = F_\text{mesh}(V_\text{template}, f_\text{pho}),
\end{equation}
where $V_\text{template}$ is the zero-pose template for audio features.

Lip-related vertices and phoneme-related features are separately encoded and concatenated to predict expressions and poses of the FLAME model by a learnable network:
\begin{equation}
\hat{\psi}, \hat{\theta} = F_\text{exp}(E_\text{lip}(V_\text{lip}), E_\text{pho}(f_\text{pho})), 
\label{eq:exp}
\end{equation}
where $E_\text{lip}$ and $E_\text{pho}$ are two feature encoders and $F_\text{exp}$ is a network that concatenates two kinds of features and predicts expressions and poses. 
Meanwhile, to address the unstable prediction caused by a single audio frame, we input neighboring frames and use attention layers in $F_\text{exp}$ to integrate multi-frame audio information.
This learning process is supervised by $\mathcal{L}_\text{exp} = \| \hat{\psi} - {\psi}  \|_2^2 + \| \hat{\theta} - {\theta}  \|_2^2$.

\noindent\textbf{Neural Video Rendering Network.}
Gaining the new driven coefficients ${\psi}$, ${\theta}$, we can send them to Eq.~\ref{eq:flame} and recalculate the geometry to get a new normal $\hat{N}$ that fits input audio.
Here we find that it is non-trivial to faithfully relate the audio signals and all face deformations (\eg head movement).
The translation network $F_\text{exp}$ will perform poorly if required to fit all poses.
Therefore, we only predict lip-related poses and directly use the sequence of lip-unrelated poses from existing videos.
In this way, we only need to regenerate lip-related areas (including cheek and chin) and blend lip-unrelated areas from the existing videos.
We first use a ResNet based local network $F_\text{local}$ to translate the newly generate normal to lip-related areas. Using eroded lip-unrelated areas from existing videos as background and the output of the first network, another blending network $F_\text{blend}$ is used to output the blended image $\hat{I}$:
\begin{equation}
\hat{I} = F_\text{blend}(F_\text{local}(\hat{N}) \odot {M}, I \odot (1 - {M}^{d})),
\label{eq:blend}
\end{equation}
where ${M}$ is the lip-related area and ${M}^{d}$ is the dilated area for the network $F_\text{blend}$ to inpaint.
This process is learned by:
\begin{equation}
\mathcal{L}^\text{local}_\text{rgb} = \| F_\text{local}(\hat{N}) \odot {M} - I \odot {M} \|_2^2, 
\end{equation}

% \vspace{-0.2in}
\begin{equation}
\mathcal{L}^\text{blend}_\text{rgb} = \| \hat{I} - I  \|_2^2, 
\end{equation}
% \vspace{-0.3in}

\begin{equation}
\mathcal{L}^\text{blend}_\text{per} = \| \text{VGG}(\hat{I}) - \text{VGG}(I)  \|_2^2, 
\end{equation}
where $\mathcal{L}^\text{blend}_\text{per}$ is the perceptual loss and  VGG represents a pretrained face VGG network~\cite{2015Deep} and returns extracted embedding features. It is only added to the blending network to generate vivid results while the local network is only supposed to generate the rough lip area.

\subsection{Reflectance Decomposition}
\label{sec:reflectance}
To enable rendering the talking portrait under novel illuminations, the reflectance and environmental lighting should be appropriately disentangled and estimated. 

\noindent\textbf{Lighting.} Following previous work ~\cite{ramamoorthi2001relationship,barron2014shape,basri2003lambertian,wang2008face,shu2017neural,zhou2019deep,hou2021towards}, the environmental lighting $L$ is represented as a 9-dimensional spherical harmonics coefficient vector. However, the lighting conditions of online talking videos are unknown, which makes it hard for inverse rendering. Inspired by Relighting4D~\cite{chen2022relighting4d}, we first initialize the lighting $L$ from the front of the human face and then treat it as a trainable parameter to optimize.

\noindent\textbf{Normal Map.} Although $\hat{N}$ provides a rough estimation of portrait geometry, its deviation from the real geometry will be amplified in the relighting. To further refine the geometry while still keeping the structure of $\hat{N}$, we use a network $F_\text{normal}$ to predict normal residual:
\begin{equation}
\delta N = F_\text{normal}(\hat{I}, \hat{N}).
\end{equation}
We add an $L_1$ regularization on $\delta N$, i.e., $\mathcal{L}_{\delta N}=\|\delta N\|_1$.
The final predicted normal $N$ is $N = \hat{N} + \delta N$.

\noindent\textbf{Shading Map.} Given the normal $N$ and lighting $L$, we can calculate the shading map $S_\text{shad}$ using a network $F_\text{shad}$ conditioned on the normal and lighting:
\begin{equation}
S_\text{shad} = F_\text{shad}(N, L),
\end{equation}

\noindent\textbf{Albedo Map.} To represent the illumination-invariant base color of the human face, we use a network $F_\text{albedo}$ to predict the albedo map $A$ from the appearance:
\begin{equation}
A = F_\text{albedo}(\hat{I}).
\end{equation}

Although there is no ground truth for albedo in our setting, it is supposed to have two physical priors: smoothness and parsimony~\cite{barron2014shape}. Smoothness requires that variation in the albedo map tends to be small and sparse. To achieve that, we use total variation regularization on the skin area:
\begin{equation}
\begin{aligned}
\mathcal{L}_{\text {smooth}}(A)&=\sum_{h=1}^{H} \sum_{w=1}^{W}\left\|\boldsymbol{\beta}_{h+1, w}-\boldsymbol{\beta}_{h, w}\right\|_2^2 \\
&+\sum_{h=1}^{H} \sum_{w=1}^{W}\left\|\boldsymbol{\beta}_{h, w+1}-\boldsymbol{\beta}_{h, w}\right\|_2^2,
\label{eq:smooth}
\end{aligned}
\end{equation}
where $\boldsymbol{\beta}_*$ are the values of albedo $A$ within the skin area.

In addition to piece-wise smoothness, the second property we expect from albedo map is parsimony, which means that the palette with which an albedo image was painted should be small. This property holds only when it is a soft constraint to make the palette sparse enough. 
As for the parsimony prior, we penalize the network by minimizing the entropy of the albedo map \cite{chen2022relighting4d}:
\begin{equation}
\mathcal{L}_{\text{parsimony}} =\mathbb{E}[-\log(p(A))],
\label{eq:parsimony}
\end{equation}
where $p(\cdot)$ is the probability density function (PDF). 
To address the difficulty of estimating the PDF of the continuous variable albedo map $A$ during training, we use Monte Carlo sampling to obtain a soft approximation of a Gaussian histogram at predefined bins for estimating the PDF of $A$.

\noindent\textbf{Specular Map.}
Prior works~\cite{he2016deep, shu2017neural} for portrait relighting, especially which require no One-Light-at-A-Time (OLAT) data, only employ simple diffuse lighting to model the human face. However, given the fact that specular phenomenons widely appear on human faces, it is key to the photorealistic rendering to model the specular effects. Therefore, we leverage Blinn-Phong model~\cite{blinn1977models} to incorporate specular component as:
\begin{equation}
R_\text{spec}\left(N, \omega_i, \omega_o\right)= \frac{s+2}{2 \pi}\left(h\left(\omega_i, \omega_o\right) \cdot N \right)^s,
\end{equation}
where $h(\omega_i, \omega_o) = \text{normalize}(\omega_i + \omega_o)$, and $s$ is the Phong exponent that controls the apparent smoothness of the surface.
Then the specular map $S_\text{spec}$ can be calculated by the accumulation of $R^\text{spec}\left(N, \omega_i, \omega_o\right)$ under illumination from different directions:
\begin{equation}
\small
S_\text{spec} = F_\text{spec}(N, L) = \sum_{\omega_i}(L(\omega_i) \odot R_\text{spec}\left(N, \omega_i, \omega_o\right)),
\end{equation}
in which $\omega_o$ is always towards the front in this paper. In experiments, we also find that the specular produced by Blinn-Phong model can never perfectly align with the real face in the video. Inspired by SunStage~\cite{wang2022sunstage}, we use another network $F_\text{cspec}$ to predict a coefficient map $C_\text{spec}$ for flexibly adjusting the final specular:
\begin{equation}
C_\text{spec} = F_\text{cspec}(\hat{I}, N).
\end{equation}

For the coefficient map, we apply a TV loss mentioned in Eq.\ref{eq:smooth} to avoid checkerboard artifacts.
Finally, we synthesize the video frame $\tilde{I}$ via image-based rendering:
\begin{equation}
F_\text{render}: \tilde{I} = A\odot(S_\text{shad}+C_\text{spec} \odot S_\text{spec}),
\end{equation}
where $\odot$ denotes the element-wise product. And the training objective is the reconstruction loss against input frames:
\begin{equation}
\mathcal{L}^\text{render}_\text{rgb} = \| \tilde{I} - I \|_2^2.
\end{equation}

\noindent\textbf{Identity-Consistent Supervision with Relighting.}
Coarse renderer $F_\text{coarse}$ synthesizes RGB pixel values according to the normal $N$. Without multi-view supervision, the face area in highlights will be regarded as raised part even if it is smooth originally, leading to artifacts during the relighting. To address this issue, we propose identity-consistent supervision with simulated multiple lighting conditions, which is performed on-the-fly during training. We assume that a well-trained face recognition network can extract similar embedding when the lighting condition varies. Therefore, after the decomposition is nearly converged, we randomly sample a new lighting condition and reinforce the identity consistency between the two rendered images with different lighting conditions:
\begin{equation}
\mathcal{L}_\text{consistent} = \| E_\text{id}(I^\text{relight}) - E_\text{id}(I) \|_2^2,
\end{equation}
where $I^\text{relight}$ is the rendering under the randomly sampled lighting, and $E_\text{id}$ is the embedding extracted by a pre-trained face recognition network~\cite{schroff2015facenet}.

\begin{figure}
    \centering
    \begin{minipage}{.45\textwidth}
        \centering
         \includegraphics[width=0.98\linewidth]{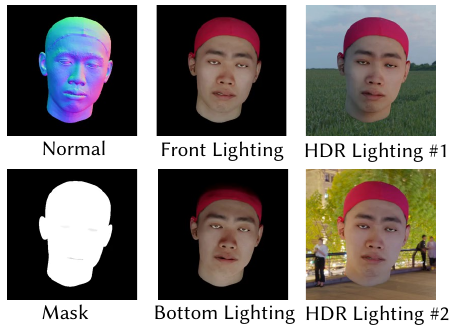}
        \caption{\textbf{Visualization of Synthetic Data.} We render $6$ sequences ($2$ minutes, $25$ fps), for each person in $10$ different lighting conditions with Cycles renderer~\cite{hess2013blender} in Blender\cite{blender}.}
  \label{fig:syn}
    \end{minipage}%
\end{figure}

\subsection{Optimization and Inference}

During the training phase, training the entire framework directly may cause the networks to learn the locally optimized results of each decomposed map, as there are no ground truths available for each component of reflectance decomposition.
Therefore, we first train networks for audio-driven synthesis to learn a rough expression- and pose-related geometry. 

Yet, there is no off-the-shelf ground truth for normal map $N$ to supervise geometry refinement. To address it, a coarse renderer $F_\text{coarse}$ is used to predict the RGB result $\hat{I}$ conditioning on normal $N$, which is supervised by image reconstruction loss. This self-supervised learning process encourages the normal map to obtain more details of surface shape, without requiring extra annotations.
After a rough normal map $N$ is gained, it is combined with an RGB portrait image as inputs of reflectance decomposition for further optimization and also stabilize the decomposition. 

The overall loss is:
\begin{equation}
\begin{aligned}
\mathcal{L} &= \lambda^\text{local}_\text{rgb} \mathcal{L}^\text{local}_\text{rgb} + \lambda^\text{blend}_\text{rgb} \mathcal{L}^\text{blend}_\text{rgb} + \lambda^\text{blend}_\text{per} \mathcal{L}^\text{blend}_\text{per} \\
& + \lambda^\text{render}_\text{rgb} \mathcal{L}^\text{render}_\text{rgb} + 
\lambda_{\delta N} 
\mathcal{L}_{\delta N} + 
\lambda_\text{consistent}
\mathcal{L}_\text{consistent} \\
& + \lambda_\text{exp}
\mathcal{L}_\text{exp} + 
\lambda_{\text{parsimony}} 
\mathcal{L}_{\text{parsimony}} + \lambda^\text{total}_\text{smooth} \mathcal{L}^\text{total}_\text{smooth},
\end{aligned}
\end{equation}
where $\lambda$’s are the weights and are set to 1, 1, 100, 1, 1, 3, 1, 0.001, and 1 respectively.
Here $\mathcal{L}^\text{total}_\text{smooth}$ is similar to Eq.~\ref{eq:smooth} but is added to all decomposed maps:
\begin{equation}
\begin{aligned}
\mathcal{L}^\text{total}_\text{smooth} &= \mathcal{L}_{\text {smooth}}(A) + \mathcal{L}_{\text {smooth}}(N) \\
& + \mathcal{L}_{\text {smooth}}(R_\text{spec}) + 
\mathcal{L}_{\text {smooth}}(C_\text{spec}).
\end{aligned}
\end{equation}

In the inference phase, new audio will drive the portrait by controlling expression and pose coefficients. Meanwhile, desired illuminations will replace the learned light $L$ of the original video to relight the whole video, thus seamlessly harmonizing with the background.

\begin{figure}[t]
% \vspace{-0.1in}
\centering
\includegraphics[width=0.90\linewidth]{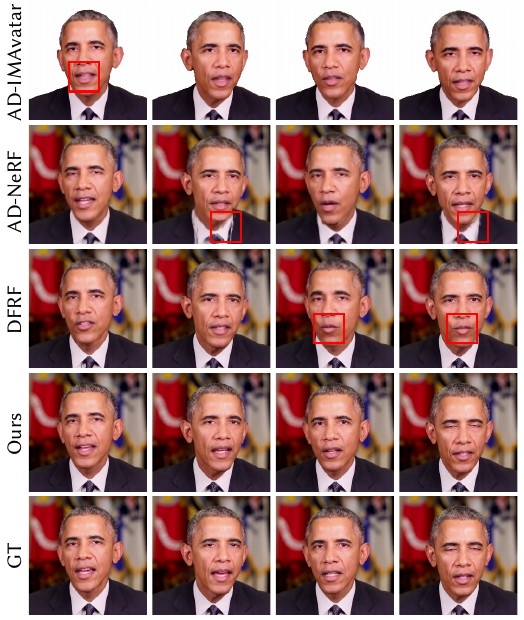}
\vspace{-0pt}
\caption{\textbf{Qualitative Comparison of Real Video Driving.} Our method successfully drives the motion of lips. Compared to AD-IMAvatar~\cite{zheng2022avatar}, AD-NeRF~\cite{guo2021ad}, and DFRF~\cite{shen2022dfrf}, our generated lip motion are closer to the ground truth (zoom in for a better view).}
\label{fig:audcompare}
% \vspace{-0.25in}
\end{figure}

\subsection{Implementation Details}

In audio-driven synthesis, lip-related feature encoder $E_\text{exp}$ and phoneme-related feature encoder $E_\text{pho}$ both use $1$D convolutional neural networks. 
Decoder $F_\text{exp}$ is also a $1$D convolutional neural network but with the self-attention mechanism~\cite{zhang2019self} to predict pose and expression coefficients through $8$ adjacent frames. 
For Local network $F_\text{local}$, we use the ResNet~\cite{he2016deep} with $6$ residual blocks. While for blending network $F_\text{blend}$, we use U-Net of depth $5$ with dilated convolutions~\cite{thies2020neural}. To gain coherent results, we adjust the mask size to leave some missing area between the generated lip area and the given background area, which will be inpainted by $F_\text{blend}$. As shown in Fig.~\ref{fig:details}, we choose the area with $80\times80$ resolution around the mouth as the Driven Area and remove the area with $120\times120$ resolution around the mouth as the Existing Area. Here we also add the lip area generated by Wave2Lip~\cite{prajwal2020lip} as the additional input of $F_\text{blend}$ to increase the performance when the input audio is far away from the audio used in training (i.e. audio from a new person).

For reflectance decomposition, we choose U-Net~\cite{isola2017image} of depth $8$ as the architecture of specular weight predicter $F_\text{cspec}$. But to gain smoother albedo maps and normal residuals, we choose ResNet~\cite{he2016deep} with $6$ residual blocks as the architecture of albedo predictor $F_\text{albedo}$ and normal residual predictor $F_\text{normal}$.

\begin{figure*}[t]
% \vspace{-0.1in}
\centering
\includegraphics[width=0.90\linewidth]{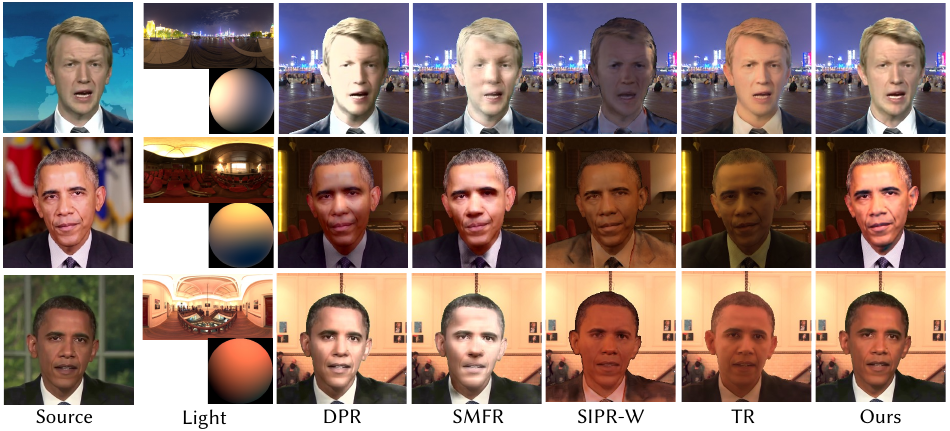}
\vspace{-0pt}
\caption{\textbf{Qualitative Comparisons of Real Video Relighting.} We compare our methods against DPR~\cite{zhou2019deep}, SMFR~\cite{hou2021towards}, SIPR-W~\cite{wang2020single} and TR~\cite{pandey2021total}.
ReliTalk renders human portraits with high-fidelity even with the complex lighting from HDR environment maps.
}
% \vspace{-0.25in}
\label{fig:lighthdr}
\end{figure*}

\begin{figure}[t]
% \vspace{-0.1in}
\centering
\includegraphics[width=0.90\linewidth]{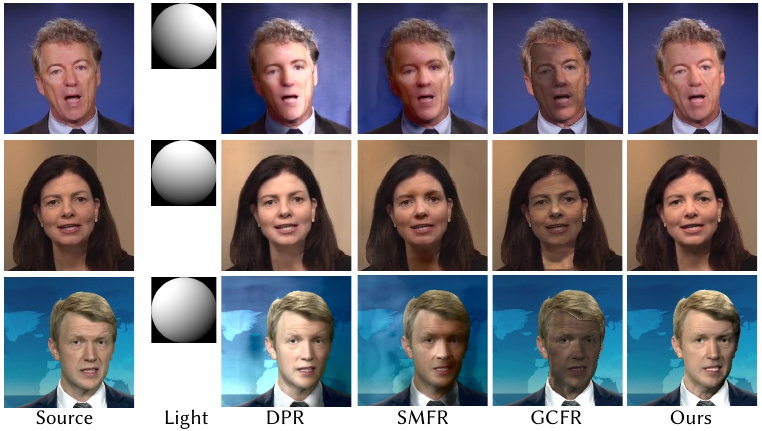}
\vspace{-0pt}
\caption{\textbf{Qualitative Comparisons Under Directional Lights.} We compare our methods against three baseline methods DPR~\cite{zhou2019deep}, SMFR~\cite{hou2021towards} and GCFR~\cite{hou2022face} for portrait relighting under directional light.
}
% \vspace{-0.25in}
\label{fig:lightdir}
\end{figure}

\section{Experiments}
\label{sec:exp}

\subsection{Dataset}

\noindent\textbf{Real Video Data.}
AD-NeRF~\cite{guo2021ad} and HDTF~\cite{zhang2021flow} collect several high-resolution talking videos in different scenes to better evaluate the generation performance. Following this practice, we choose celebrity videos whose protagonists are news anchors, entrepreneurs, or presidents from YouTube as our real video set.
We collect $8$ public videos with an average length of $3$ minutes from $7$ identities. We split each video with around $80\%$ frames for training and $20\%$ frames for evaluation. These videos are all available online and we will provide corresponding source links for reproduction purposes.

\noindent\textbf{Synthetic Video Data.}
Talking videos with ground-truth illuminations are not available from online collections.
To evaluate our relighting algorithm quantitatively, we synthesize some talking videos with the same motion sequence but different lighting conditions within the modern graphic pipeline, as shown in Fig.~\ref{fig:syn}.
Specifically, we render $6$ sequences ($2$ minutes, $25$ fps), for each person in $10$ different lighting conditions with Cycles renderer~\cite{hess2013blender} in Blender\cite{blender}, a photorealistic ray-tracing renderer. All mesh models, textures, and displacement maps are released by FaceScape~\cite{yang2020facescape, zhu2021facescape}.  We combine displacement maps and textures in a physically-based skin material featuring sub-surface scattering~\cite{christensen2015approximate} for photo-realistic rendering.  We drive our head models with expression coefficients and head rotation angles estimated from our own recorded talking videos.

\subsection{Evaluation Metrics}
For evaluation metrics, we report peak signal-to-noise ratio (PSNR), structural similarity (SSIM), and perceptual similarity (LPIPS)~\cite{zhang2018perceptual} to measure the quality of generation results. For datasets, we collect $8$ talking portrait videos from YouTube with an average length of around $3$ minutes as our real video set (most are used in AD-NeRF~\cite{guo2021ad} or HDTF~\cite{zhang2021flow}) and additionally render synthetic videos of $6$ persons with an average length of around $2$ minutes for quantitative comparison. More details are introduced in our supplementary materials.
To measure audio-driven accuracy, we further use SyncNet (confidence) ~\cite{chung2017out} to measure the audio-driven synchronization.

\begin{table*}[t]
% \vspace{-0pt}
\begin{center}
\caption{\textbf{Quantitative Results of Audio-Driven Real Videos.} Our method significantly outperforms all baselines in terms of all metrics.}
\vspace{-0pt}
\label{table:real_video}
\scalebox{0.98}{\begin{tabular}{l|c|c|c|c}
\hline\noalign{\smallskip}
Methods & PSNR $\uparrow$ & SSIM $\uparrow$ & LPIPS $\downarrow$ & SyncNet $\uparrow$ \\
\noalign{\smallskip}
\hline
\noalign{\smallskip}
AD-IMAvatar~\cite{zheng2022avatar} & 25.0625 & 0.8885 & 0.0538 & 2.5428   \\
AD-NeRF~\cite{guo2021ad} & 25.6916(29.8458) & 0.9219(0.9750) & 0.1165(0.0594) & 3.8616   \\
DFRF~\cite{shen2022dfrf}  & 33.2088(33.9563) & 0.9665(0.9834) & 0.1178(0.0616) & 3.7190   \\
Ours & \textbf{37.6645(37.9082)} & \textbf{0.9892(0.9931)} & \textbf{0.0028(0.0029)} & \textbf{5.5343}   \\
\hline
Ground Truth & - & 1.000 & 0.000 &  7.7218   \\
\hline
\end{tabular}}
% \vspace{-0.3in}
\end{center}
\end{table*}
\setlength{\tabcolsep}{1.4pt}

\subsection{Qualitative Comparison}

Currently, there is no unified framework for relightable audio-driven talking portrait generation. Therefore, we first compare our method with audio animation methods and relighting methods separately, then trivially combine two existing frameworks as the baseline of relightable audio-driven talking portrait generation.

\noindent\textbf{Comparison on Audio-Driven Talk.}
In this work, we focus on personalized animation, which only uses one video for training.
We choose two representative personalized audio animation methods, AD-NeRF\cite{guo2021ad}, and DFRF~\cite{shen2022dfrf}. We also modify the FLAME-based method IMAvatar~\cite{zheng2022avatar} to gain a simple audio-driven version, AD-IMAvatar as an additional baseline.
In Fig.~\ref{fig:audcompare}, AD-IMAvatar only generates coarse talking portraits with blurry teeth areas. 
And AD-NeRF is prone to generate artifacts at the junction of the neck and head. 
Compared to the results of AD-NeRF and DFRF, the motion of our generated lips is closer to the ground truth.  
Notably, our framework succeeds to generate clear teeth areas.

\noindent\textbf{Comparison on Relighting.}
In this paper, we compare our relighting performance with five advanced methods. DPR~\cite{zhou2019deep}, SMFR~\cite{hou2021towards} and GCFR~\cite{hou2022face} are trained on publicly available data and release their models. Since nearly none of One-Light-at-A-Time (OLAT) based methods release their code or models. We requested the authors to inference their models on our provided inputs (SIPR-W~\cite{wang2020single} and TR~\cite{pandey2021total}), and take results for comparisons.

As presented in Fig.~\ref{fig:lightdir}, although both DPR and SMFR are able to reflect given lighting conditions on generated images when a sample directional light is given, their generated portraits lack the special texture of a real human face. 
This is mainly because they do not account for model specular, which is a significant and noticeable feature of the human face.
Meanwhile, the recent method GCFR is easy to produce unnatural shadows.
In contrast, ReliTalk renders realistic human portraits with reserved facial details.

Additionally, when complex lighting optimized from HDR images is used (as shown in Fig.~\ref{fig:lighthdr}), DPR and SMFR tend to produce unsatisfactory results, many of which are not relevant to the given lighting conditions.
And SMFR even fails to reconstruct some faces.
SIPR-W will generate some relighted results whose color is similar to the background but can not reflect the varied lighting on the face.
Although TR succeeds to generate some vivid relighted results, it loses some facial details and also mildly hurt the original identity.
However, our framework performs well on both types of lighting and successfully renders the specular texture of the human face. This enables our generated avatar to blend in seamlessly with various backgrounds, as long as HDR data of the background is available, by matching the shading and lighting of the avatar to that of the background.

\subsection{Quantitative Comparison}

\noindent\textbf{Evaluation on Audio-Driven Talk.}
As shown in Table~\ref{table:real_video}, we compare our method with AD-IMAvatar, AD-NeRF, and DFRF. 
Among those baselines, DFRF achieves comparable performance in PSNR, SSIM, and LPIPS, while its confidence in SyncNet is slightly lower than AD-NeRF. However, our method significantly outperforms all baselines in terms of all metrics.

Although our method uses some portrait area from existing frames, both AD-NeRF and DFRF use pose parameters from the existing sequence. In this way, DFRF only generates the remaining face area with the neck and collar given. Therefore for a fair comparison, we recalculate PSNR, SSIM, and LPIP purely in the driven area ($120\times120$ resolution). As shown in the brackets of Table~\ref{table:real_video}, our method still significantly outperforms all baselines in terms of all metrics. 

\noindent\textbf{Evaluation on Synthetic Relighting Dataset.}
As shown in Table~\ref{table:syn_video}, we achieve the highest PSNR and SSIM on the synthetic dataset. And they are significantly higher than the other two, which indicates our generated images is closer to the ground truth. Meanwhile, the lowest LPIPS also illustrates that our results have the highest perceptual quality.
It is notable that SMFR almost fails in our synthetic video dataset, which is perhaps caused by the distribution gap between our synthesized video data and real data. 
Our ReliTalk outperforms DPR and SMFR both qualitatively and quantitatively. According to the analysis of practicality, they need a pre-collect face image dataset to train the network. Our ReliTalk outperforms DPR and SMFR both qualitatively and quantitatively. According to the analysis of practicality, they need a pre-collect face image dataset to train the network. At the inference stage, a new lighting or a new portrait which is out of training distribution will significantly hurt the performance. Although our method cannot handle all persons within one model, our method is still practical as the training data, a short talking portrait video, is easy to obtain. 
While our method may not be able to handle all persons within a single model, it is still practical because the training data, a short talking portrait video, is readily available and easy to obtain.

\begin{table}[t]
% \vspace{-8pt}
\begin{center}
\caption{\textbf{Quantitative Results of Synthetic Video Relighting.} Our method achieves the highest PSNR and SSIM on the synthetic dataset.}
\vspace{-0pt}
\label{table:syn_video}
\scalebox{0.98}{\begin{tabular}{l|c|c|c}
\hline\noalign{\smallskip}
Methods & PSNR $\uparrow$ & SSIM $\uparrow$ & LPIPS $\downarrow$ \\
\noalign{\smallskip}
\hline
\noalign{\smallskip}
DPR~\cite{zhou2019deep}  & 18.1899  & 0.9093  & 0.0668  \\
SMFR~\cite{hou2021towards} & 15.9565  & 0.8003  & 0.3358  \\
\hline
Ours & \textbf{ 22.8152 }  & \textbf{ 0.9435 }  & \textbf{ 0.0326 }  \\
\hline
\end{tabular}}
\end{center}
% \vspace{-0.3in}
\end{table}
\setlength{\tabcolsep}{1.4pt}

\begin{table}[t]
% \vspace{-8pt}
\begin{center}
\caption{\textbf{Ablation results of Mesh-Aware Guidance.} Our mesh-aware guidance improves prediction accuracy significantly.}
\vspace{-0pt}
\label{table:abl_mesh}
\scalebox{0.98}{\begin{tabular}{l|c|c}
\hline\noalign{\smallskip}
Methods & PSNR $\uparrow$ & SSIM $\uparrow$  \\
\noalign{\smallskip}
\hline
\noalign{\smallskip}
Audio Only & 29.6147   & 0.9511     \\
Mesh Only & 33.8643   & 0.9790     \\
\hline
Audio + Mesh (Ours) & \textbf{ 34.1099 }   & \textbf{ 0.9802 }     \\
\hline
\end{tabular}}
\end{center}
% \vspace{-0.3in}
\end{table}
\setlength{\tabcolsep}{1.4pt}

\subsection{Ablation of Core Modules}

\noindent\textbf{Effects of Mesh-Aware Guidance.}
Mesh-aware guidance is used to assist lip-syncing. We use a model pre-trained~\cite{8954000}
to gain lip-related meshes as additional guidance. Phoneme-related features and lip-related meshes are separately encoded and then concatenated to generate pose and expression coefficients.
As shown in Table~\ref{table:abl_mesh}, mesh-aware guidance improves prediction accuracy significantly.
In addition, we find that the improvement is significant (PSNR increases from $29.5445$ to $34.8186 $) when the training video is too short to cover enough phonemes ($2450$ training frames). The improvement is mild for the long video ($6500$ training frames).
This implies that mesh-aware features offer the network approximate information about the movements of the lips, such as opening and closing, even when the input audio is significantly different from the audio used during training.

\begin{figure}[t]
% \vspace{-0.1in}
\centering
\includegraphics[width=0.90\linewidth]{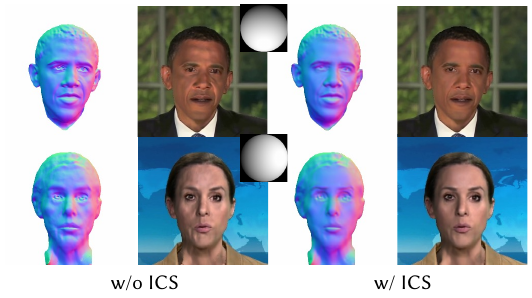}
\vspace{-0pt}
\caption{\textbf{Finer Normal under Identity-Consistent Supervision with Relighting.} Without ICS, the normal map estimation may contain severe artifacts which causes unrealistic rendering given novel illuminations.
}
\label{fig:abl_ics}
\vspace{-0pt}
\end{figure}

\noindent\textbf{Effects of Identity-Consistent Supervision.}
Identity-consistent supervision with relighting is employed to lessen the influence of lacking multi-view information.
We visualize the effects in Fig.~\ref{fig:abl_ics}. Instead of a well-structured normal, the network prefers to generate an irregular one whose surface varies with the change of color on the face because it is an easier mapping for the coarse render.
However, those irregular areas will be very significant when a different lighting is given (left of Fig.~\ref{fig:abl_ics}).
After adding identity-consistent supervision, this weird face is hard to be recognized as the same person, urging the network to produce a well-structured normal which can gain reasonable relighting results under lighting with various directions (right of Fig.~\ref{fig:abl_ics}).
To quantitatively evaluate the improvements of identity-consistent supervision, we calculate metrics in the image space of the normal map. Here we propose $\text{PSNR}_\text{grad}$, which calculates the 2D gradient in the image space, to jointly measure the normal quality. 
As shown in Table~\ref{table:abl_ics}, although normal refined by ICS does not gain a higher PSNR, its $\text{PSNR}_\text{grad}$ is significantly higher, which means that it owns a better shape surface. Higher SSIM also proves the effectiveness of our method. 

\begin{table}[t]
% \vspace{-8pt}
\begin{center}
\caption{\textbf{Ablation Results of Identity-Consistent Supervision.} Without identity-consistent supervision, the estimated normal map will become noisy and contain more artifacts, indicated by a higher error in the gradient map.}
\vspace{-0pt}
\label{table:abl_ics}
\scalebox{0.98}{\begin{tabular}{l|c|c|c}
\hline\noalign{\smallskip}
Methods & PSNR $\uparrow$ & $\text{PSNR}_\text{grad}$ $\uparrow$ & SSIM $\uparrow$ \\
\noalign{\smallskip}
\hline
\noalign{\smallskip}
w/o ICS & 21.7141  & 21.9828   & 0.9060  \\
w ICS  & 21.5835  & 23.5441  & 0.9203  \\
\hline
\end{tabular}}
\end{center}
% \vspace{-0.3in}
\end{table}
\setlength{\tabcolsep}{1.4pt}

\begin{figure*}[t]
% \vspace{-0.1in}
\centering
\includegraphics[width=0.98\linewidth]{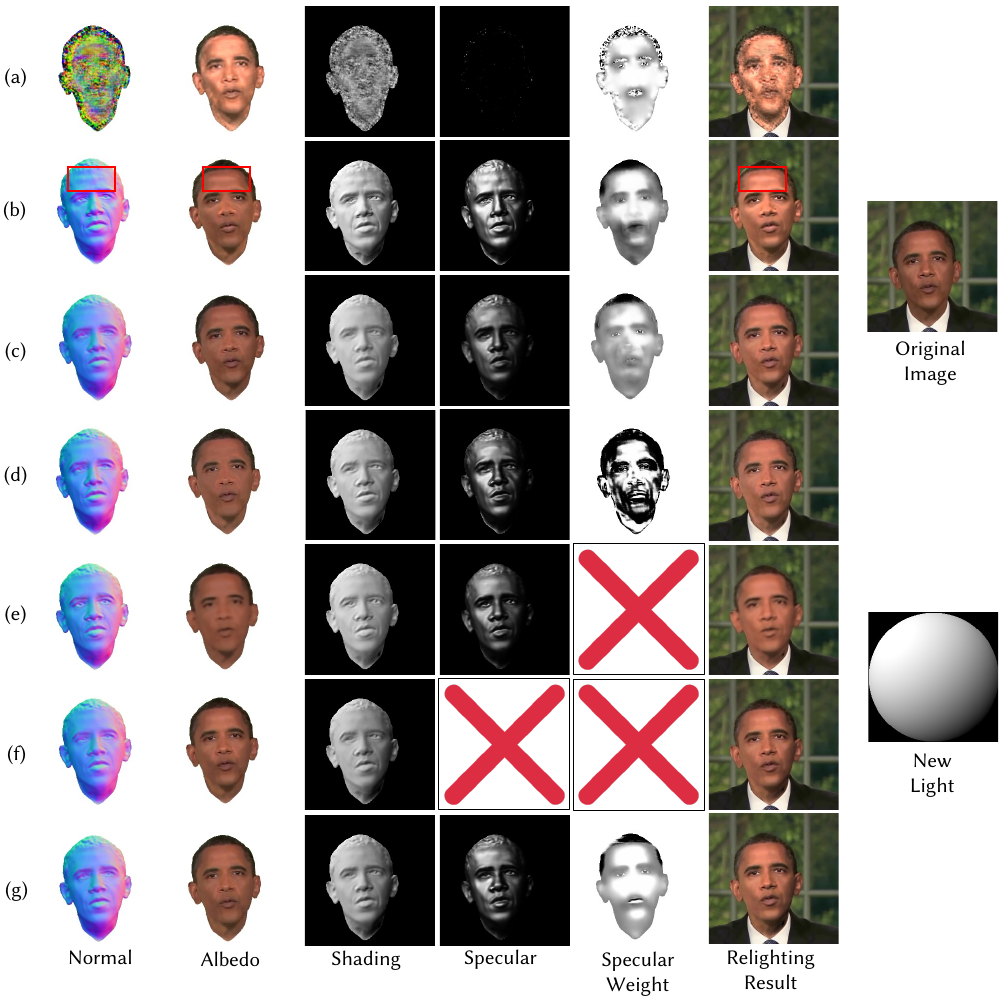}
% \vspace{-0pt}
\caption{\textbf{Ablation of Reflectance Decomposition.} \textbf{(a)} Without initial normal, \textbf{(b)} without normal residual, \textbf{(c)} without parsimony prior, \textbf{(d)} without smoothness constraints, \textbf{(e)} without specular weight, \textbf{(f)} without specular map, \textbf{(g)} our final method. Our final method gains the most vivid relighting results.
}
% \vspace{-0.25in}
\label{fig:lightabl}
\end{figure*}

\subsection{Ablation of Reflectance Decomposition}

To get rid of leveraging knowledge from expensive capturing data (\ie Light Stage~\cite{lightstage}), we decompose the human portrait into corresponding maps from monocular videos through some careful designs.

\noindent\textbf{Initial Normal.} Different from previous audio-driven generation methods only generate the final portrait image, our audio-to-geometry also provides a good initial normal estimation. As shown in row \textbf{(a)} of Fig.~\ref{fig:lightabl}, the reflectance decomposition can not converge without initial normal estimation because of lacking the constraints for the normal map.

\noindent\textbf{Normal Residual.} Initial normal is not accurate because we do not have either the ground truth of the normal map or multi-view information of the portrait. Those irregular areas will be very significant when new lighting is given (row \textbf{(b)} of Fig.~\ref{fig:lightabl}).

\noindent\textbf{Parsimony Prior.} Parsimony means that the palette with which an albedo image was painted should be small. Without parsimony prior, albedo will contain more details while normal details are reduced, which is obviously reflected in the shading map of row \textbf{(c)} (Fig.~\ref{fig:lightabl}). Therefore, the final relighting result is not such vivid.

\noindent\textbf{Smoothness Constraints.} With smoothness constraints, some decomposition maps may overfit training data. As shown in row \textbf{(d)} of Fig.~\ref{fig:lightabl}, the predicted specular weight is chaotic, reducing the vividness of the relighting result.

\noindent\textbf{Specular Weight.} We use a network $F_\text{cspec}$ to predict a specular weight $C_\text{spec}$ for flexibly adjusting the final specular. As shown in row \textbf{(e)} of Fig.~\ref{fig:lightabl}, the relighting result is blurred without this design.

\noindent\textbf{Specular Map.} Prior works~\cite{he2016deep, shu2017neural} for portrait relighting only employ simple diffuse lighting to model the human face. However, ignoring specular phenomenons that widely appear on human faces, the final rendering result is less photo-realistic (row \textbf{(f)} of Fig.~\ref{fig:lightabl}).

\section{Conclusion}

We propose \textbf{ReliTalk} a novel framework for relightable audio-driven talking portrait generation which only requires an easily accessible single monocular portrait video as input, while previous light-stage-based methods are not publicly available for data or code.
Our method decomposes the human portrait for the well-disentangled geometry and reflectance, which is also expression- and pose-related.
During the inference, we use audio from the user to drive the human portrait by controlling expression and pose coefficients, then render it with any desired illuminations, seamlessly harmonizing with the background.

However, there are still some limitations of our designed relighting model.
1) We only consider one-bounce direct environment light, and thus our method cannot handle furry appearances, such as beards and long hair. 2) Our method assumes the appearance of human faces does not change throughout the entire video. Therefore, actions like wearing glasses or putting on hats may change the appearance would cause inaccurate estimation of reflectance.

In the future, we want to design a more realistic physical model that can take into account various complex lighting conditions.

\noindent\textbf{Societal Impacts}
Our code is released for better promotion. 
Therefore, users only need to input a talking video of the target person and then are able to freely generate a talking portrait with desired audio and background.
Although it increases the risk of forged videos, our approach also provides a new type of forged samples for researchers to improve defense methods.

\section{Acknowledgements}

This research is supported by the National Research Foundation, Singapore under its AI Singapore Programme (AISG Award No: AISG2-PhD-2022-01-035T), NTU NAP, MOE AcRF Tier 1 (2021-T1-001-088), and
under the RIE2020 Industry Alignment Fund – Industry Collaboration Projects (IAF-ICP) Funding Initiative, as well as cash and
in-kind contribution from the industry partner(s).

\section{Data Availability Statement}

Video data and pre-trained models used in this paper are available online. We provide corresponding source links for reproduction purposes in the \textbf{ReliTalk} repository \href{https://github.com/arthur-qiu/ReliTalk}{https://github.com/arthur-qiu/ReliTalk}.


\bibliographystyle{spbasic}      % basic style, author-year citations
\bibliography{ref}   % name your BibTeX data base

\end{document}